\documentclass[letterpaper, 10pt, conference]{ieeeconf}

\IEEEoverridecommandlockouts   
\usepackage{cite}

\usepackage{graphicx}
\usepackage{epsfig}

\usepackage{amsmath,amsfonts}
\usepackage{amssymb}

\usepackage{booktabs}
\usepackage{multirow}
\usepackage{multicol}
\usepackage{array}

\usepackage{algorithmic}
\usepackage[ruled,vlined]{algorithm2e}

\usepackage[caption=false,font=normalsize,labelfont=sf,textfont=sf]{subfig}

\usepackage{textcomp}
\usepackage{stfloats}
\usepackage{url}
\usepackage{verbatim}
\usepackage{caption}
\usepackage{balance}

\usepackage[dvipsnames, svgnames, x11names]{xcolor}
\usepackage{colortbl}

\usepackage{pifont}

\definecolor{convcolor}{HTML}{412F8A}
\definecolor{resnetcolor}{HTML}{8DA0CB}
\definecolor{vitcolor}{HTML}{fc8e62}

\newcommand{\methodname}{{WAM}\xspace}

\usepackage{booktabs}                                   
\usepackage{multirow}                                   
\usepackage{makecell}                                   
\usepackage{tablefootnote}                              
\usepackage[symbol]{footmisc}                           
\usepackage{amsmath,amssymb}                            
\usepackage{xcolor}                                     
\let\labelindent\relax              
\usepackage{enumitem}                                   
\usepackage{subcaption}                                 
\usepackage{stfloats}                                   
\usepackage[misc]{ifsym}     
\usepackage{hyperref}   

\newcommand{\eat}[1]{}                                  

\title{\LARGE \bf Enhancing Policy Learning with World-Action Model}

\author{Yuci Han$^{1}$ and Alper Yilmaz$^{1}$%
\thanks{$^{1}$Yuci Han and Alper Yilmaz are with the Photogrammetry and
Computer Vision Lab, The Ohio State University, Columbus, OH 43210, USA.
{\tt\small \{han.1489, yilmaz.15\}@osu.edu}}%
}

\begin{document}

\maketitle
\thispagestyle{empty}
\pagestyle{empty}

\begin{abstract}
This paper presents the World-Action Model (\methodname), an
action-regularized world model that jointly reasons over future visual
observations and the actions that drive state transitions. Unlike
conventional world models trained solely via image prediction,
\methodname\ incorporates an inverse dynamics objective into DreamerV2
that predicts actions from latent state transitions, encouraging the
learned representations to capture action-relevant structure critical for
downstream control. We evaluate \methodname\ on enhancing policy learning
across eight manipulation tasks from the CALVIN benchmark. We first
pretrain a diffusion policy via behavioral cloning on world model latents,
then refine it with model-based PPO inside the frozen world model. Without
modifying the policy architecture or training procedure, \methodname\
improves average behavioral cloning success from 59.4\% to 71.2\% over
DreamerV2 and DiWA baselines. After PPO fine-tuning, \methodname\
achieves 92.8\% average success versus 79.8\% for the baseline, with two
tasks reaching 100\%, using $8.7\times$ fewer training steps.
\end{abstract}


\section{Introduction}

\vspace{-6pt}
\begin{figure*}[t]
    \centering
    \includegraphics[width=1\textwidth]{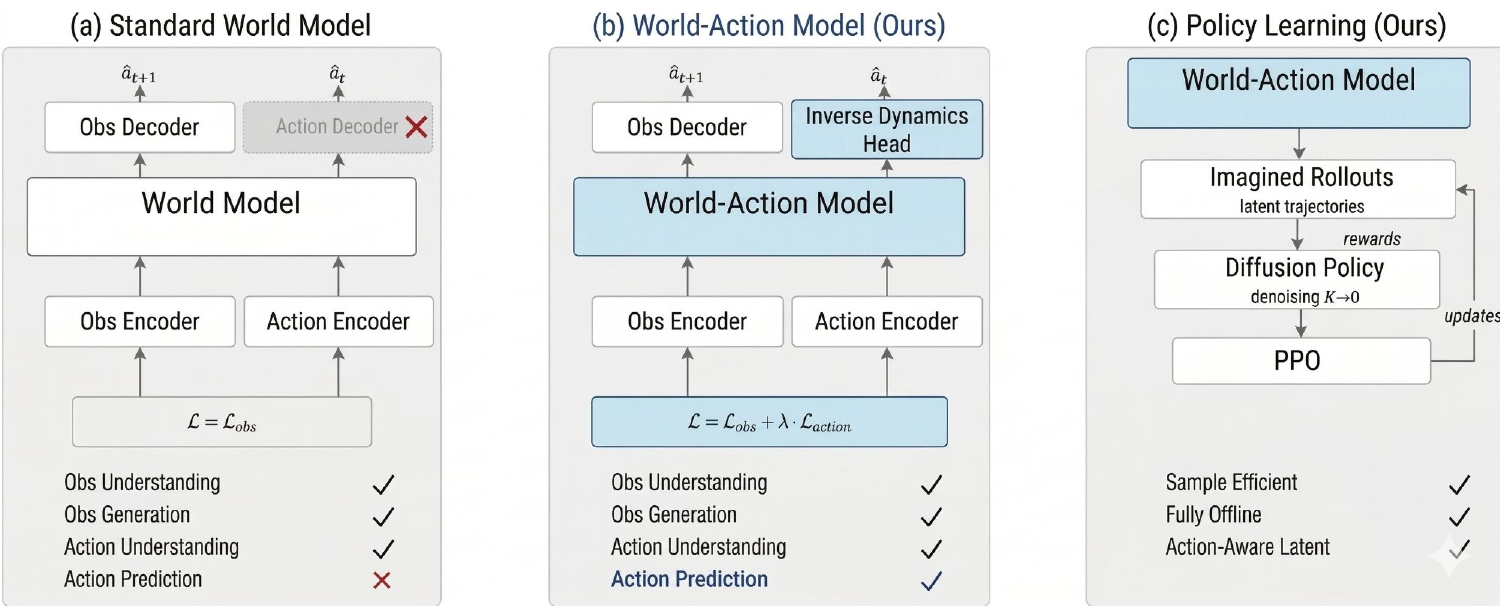}
    \caption{(a) Standard world models predict only future observations, treating actions solely as conditioning inputs. (b) Our World-Action Model adds an inverse dynamics head that jointly predicts observations and actions during training. (c) The action-aware world model serves as a learned simulator for offline policy fine-tuning via PPO.}
    \label{fig:fig1}
    \vspace{-0.5cm}
\end{figure*}

World models have become a critical tool for learning environment dynamics in robotics, allowing agents to plan and refine behaviors through imagined rollouts rather than expensive real-world interactions~\cite{ha2018world, hafner2019planet, hafner2019dreamer, hafner2020dreamerv2, hafner2023dreamerv3}. These models compress high-dimensional observations into compact latent spaces that capture the temporal and physical structures of environments, enabling sample-efficient policy learning through imagined experiences in latent space. Recent work has shown that policies trained entirely within learned world models can reach strong performance on robotic manipulation benchmarks~\cite{demoss2023ditto, nematollahi2025lumos, chandra2025diwa}, and in some cases transfer directly to physical hardware without further fine-tuning~\cite{wu2022daydreamer}.

However, conventional world models are trained solely to predict future observations~\cite{ha2018world, hafner2019planet, hafner2019dreamer, hafner2020dreamerv2, wu2022daydreamer, nematollahi2025lumos}. They forecast future visual states conditioned on past observations and actions, but never explicitly model the actions themselves. This asymmetry limits the quality of learned representations. The world model encodes each observation into a latent state $z_t$, which serves as the direct input to the downstream diffusion policy in model-based pipelines such as DiWA~\cite{chandra2025diwa}---yet $z_t$ is optimized only for pixel prediction and KL regularization, with no explicit pressure to encode action-relevant structure. Recent work on unified action-world models~\cite{cen2025worldvla} has begun to address this gap by jointly generating actions and images within a single autoregressive architecture, but these approaches rely on large foundation models and fundamentally redesign the architecture rather than improving the representations learned by existing world models.

We take a different and complementary approach: rather than redesigning the architecture, we augment the training objective of DreamerV2~\cite{hafner2020dreamerv2} with an inverse dynamics head that predicts actions from consecutive encoder embeddings, regularizing the learned representations toward action-relevant structure. We argue that requiring the model to predict the action that caused a given state transition forces the encoder to capture fine-grained information about how the environment responds to agent behavior. This information is critical for downstream policy learning but is overlooked when the training signal comes only from observation reconstruction. This idea draws on inverse dynamics models in self-supervised representation learning~\cite{pathak2017curiosity, baker2022vpt}, where predicting actions between consecutive states yields representations focused on controllable aspects of the environment while filtering out task-irrelevant distractors.

We present the World-Action Model (\methodname) and evaluate it on enhancing policy learning across eight manipulation tasks from the CALVIN benchmark~\cite{mees2022calvin}. We first pretrain a diffusion policy via behavioral cloning on world model latents, then refine it with model-based PPO using the frozen world model as a simulator. Compared to the DiWA baseline~\cite{chandra2025diwa}, \methodname\ improves average behavioral cloning success from 59.4\% to 71.2\% and achieves 92.8\% average success after PPO fine-tuning versus 79.8\%, with two tasks reaching 100\%, using $8.7\times$ fewer training steps.

Our main contributions are:
\begin{itemize}
    \item We present \methodname, a lightweight extension of DreamerV2 that augments the training objective with an inverse dynamics head, explicitly regularizing latent representations toward action-relevant structure.
    \item We demonstrate that action regularization improves world model generation quality on the CALVIN benchmark. \methodname\ matches or exceeds DreamerV2~\cite{hafner2020dreamerv2} on LPIPS, PSNR, SSIM, and FVD with fewer training steps.
    \item We show that the improved representations enhance downstream policy learning, outperforming the DiWA baseline on both behavioral cloning and PPO fine-tuning across all eight tasks on the CALVIN benchmark.
\end{itemize}
\section{Related Works}

\subsection{World Models} World models learn a predictive representation of environment dynamics, enabling agents to optimize policies through imagined trajectories without requiring additional real-world data collection. Ha and Schmidhuber~\cite{ha2018world} introduced an early approach encoding observations into a latent space with an RNN for dynamics modeling. PlaNet~\cite{hafner2019planet} proposed the Recurrent State-Space Model (RSSM), combining deterministic and stochastic components for long-horizon prediction. The Dreamer series~\cite{hafner2019dreamer, hafner2020dreamerv2, hafner2023dreamerv3} progressively extended RSSM-based latent imagination for behavior learning. DreamerV2 replaced Gaussian latents with categorical representations, while DreamerV3 introduced robustness techniques that allowed a single set of hyperparameters to work across diverse domains. However, these world models are trained purely as observation predictors without explicitly reasoning about the actions that drive state transitions. Our work introduces action regularization into the world model, jointly predicting both actions and future observations to better capture environment dynamics.

\subsection{World Models for Policy Learning} Beyond modeling dynamics, world models can be directly used as complementary environments for policy optimization, where agents learn behaviors through imagined rollouts within the learned model. The Dreamer series~\cite{hafner2019dreamer, hafner2020dreamerv2, hafner2023dreamerv3} first demonstrated that actor-critic policies can be trained entirely through imagined latent rollouts, and DayDreamer~\cite{wu2022daydreamer} demonstrated this paradigm on physical robots. In offline settings, DITTO~\cite{demoss2023ditto} proposed imitation learning by optimizing a distance metric in a frozen world model's latent space, and LUMOS~\cite{nematollahi2025lumos} extended this to language-conditioned multi-task learning. DiWA~\cite{chandra2025diwa} fine-tuned diffusion policies via PPO within a learned world model, achieving sample efficiency orders of magnitude beyond model-free baselines. More recently, WMPO~\cite{zhu2025wmpo} introduced pixel-based world models for on-policy GRPO optimization of Vision-Language-Action models, and World4RL~\cite{jiang2025world4rl} employed diffusion-based world models for end-to-end policy refinement. While these works confirm the effectiveness of world-model-based policy learning, they rely on world models trained solely for observation prediction. Our approach enriches the world model with action prediction, providing a stronger training signal for downstream policy optimization.

\section{Method}
\label{sec:method}
\subsection{World-Action Model}

We propose to regularize the world model with an inverse dynamics objective, producing representations that are simultaneously predictive of visual dynamics and aware of the actions driving state transitions.

\subsubsection{\textbf{World Model Backbone}} WAM builds on the RSSM architecture from
DreamerV2~\cite{hafner2020dreamerv2}. A dual-stream CNN encoder processes
static and gripper camera images, fusing them with proprioceptive state to
produce embeddings $e_t \in \mathbb{R}^{1554}$. The RSSM models latent
dynamics through:
\begin{align}
    h_t &= f_\phi(h_{t-1}, z_{t-1}, a_{t-1})
    \label{eq:gru} \\
    z_t &\sim q_\phi(z_t \mid h_t, e_t)
    \label{eq:posterior} \\
    \hat{z}_t &\sim p_\phi(z_t \mid h_t)
    \label{eq:prior} \\
    \hat{o}_t &\sim p_\phi(o_t \mid f_t),
    \label{eq:decoder}
\end{align}
where $h_t$ is the deterministic recurrent state, $z_t$ is a stochastic
categorical variable (32$\times$32), and $f_t = [h_t; z_t] \in \mathbb{R}^{2048}$
is the combined latent feature used for decoding and policy learning.

\subsubsection{\textbf{Action-Regularized World Model}} 
\begin{figure}[t]
\centering
\includegraphics[width=0.5\textwidth]{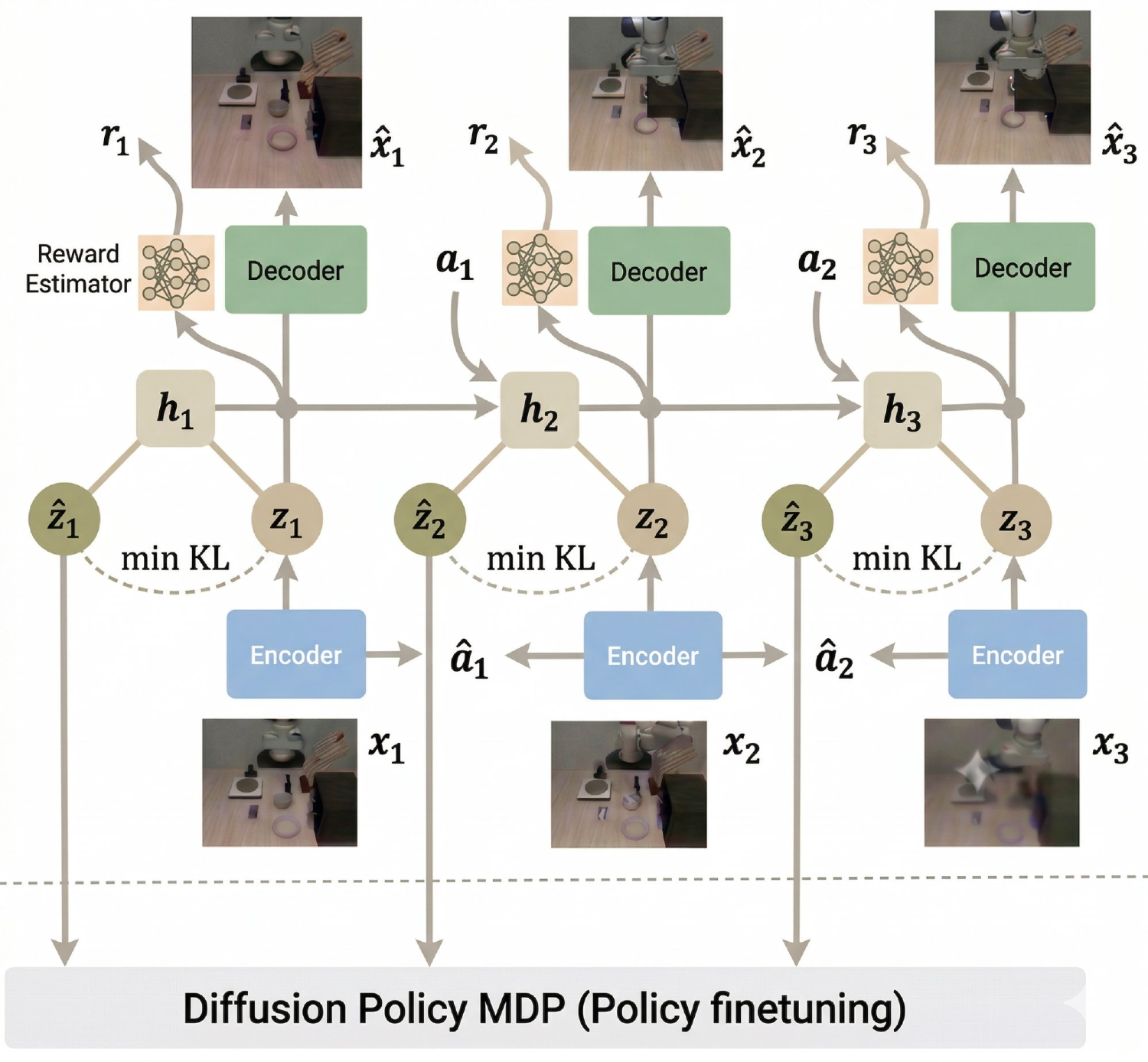}
\caption{\textbf{WAM architecture.} Observations $x_t$ are encoded and produce posterior $z_t$, which is regularized toward the prior $\hat{z}_t$ via KL divergence. The inverse dynamics head predicts actions $\hat{a}_t$ from consecutive encoder embeddings, cascading action-aware structure through the posterior to the prior. The decoder reconstructs observations $\hat{x}_t$ and a reward estimator provides task-completion signals for policy fine-tuning.}
\label{fig:rssm_unrolled}
\end{figure}

The observation-only objective
(Eq.~\ref{eq:decoder}) produces features optimized for pixel reconstruction,
potentially discarding action-relevant information. WAM addresses this by
augmenting the world model with an inverse dynamics component, yielding a
unified model $\mathcal{M}_\theta$ with two pathways:
\begin{equation}
    \hat{o}_t = \mathcal{M}_\theta^{\text{world}}(o_{1:t},\, a_{1:t-1}),
    \quad
    \hat{a}_t = \mathcal{M}_\theta^{\text{action}}(e_t,\, e_{t+1}),
    \label{eq:wam}
\end{equation}
where $\mathcal{M}_\theta^{\text{world}}$ denotes the RSSM pathway
(Eqs.~\ref{eq:gru}--\ref{eq:decoder}) and $\mathcal{M}_\theta^{\text{action}}$
is an inverse dynamics head:
\begin{equation}
    \hat{a}_t = \psi([e_t;\, e_{t+1}]),
    \label{eq:inv_dyn}
\end{equation}
with $\psi$ a three-layer MLP and $[\cdot\,;\,\cdot]$ denoting concatenation.
A critical design in WAM is the cascading effect created by the action
head through the full model. We operate on encoder embeddings $e_t$ rather than
RSSM features $f_t$, since $f_t$ receives $a_{t-1}$ through the GRU
(Eq.~\ref{eq:gru}) and would make action prediction trivially solvable. By
regularizing the encoder directly, the action-aware structure in $e_t$ cascades
forward: it shapes the posterior $z_t \sim q_\phi(z_t \mid h_t, e_t)$, which
the KL loss propagates to the prior $\hat{z}_t \sim p_\phi(z_t \mid h_t)$,
and ultimately reaches the diffusion policy through imagined rollouts generated
from $\hat{z}_t$. This chain ensures that the action prediction signal at the
encoder level translates into more informative features for downstream control. 
As illustrated in Fig.~\ref{fig:rssm_unrolled}, the action-aware structure cascades 
from the encoder through the posterior to the prior across timesteps.

\vspace{-6pt}

\textbf{Training Objective.} \methodname is trained end-to-end by minimizing:
\begin{equation}
    \mathcal{L}_{\text{\methodname}} = \lambda_{\text{KL}}\,\mathcal{L}_{\text{KL}}
    + \lambda_{\text{img}}\,\mathcal{L}_{\text{recon}}
    + \lambda_{\text{act}}\,\mathcal{L}_{\text{action}},
    \label{eq:loss}
\end{equation}
where $\mathcal{L}_{\text{KL}} = \text{KL}\big[q_\phi(z_t \mid h_t, e_t)
\,\|\, p_\phi(z_t \mid h_t)\big]$,
$\mathcal{L}_{\text{recon}} = \|o_t - \hat{o}_t\|_2^2$, and
$\mathcal{L}_{\text{action}} = \|\hat{a}_t - a_t\|_1$. We carefully tune the loss coefficients to balance the trade-off between reconstruction quality and action prediction, ensuring that the encoder learns representations that serve both objectives effectively.

\subsection{Enhancing Downstream Policy Learning with \methodname}

After world model training, we leverage WAM to enhance downstream policy
learning. The frozen \methodname defines a latent space MDP
$\mathcal{M}_{\text{wm}} = (\mathcal{Z}, \mathcal{A}, P_\phi, R_\psi,
\gamma)$, where $\mathcal{Z}$ is WAM's latent space, $P_\phi$ the learned
dynamics, and $R_\psi$ a learned reward classifier. We extract features
$f_t \in \mathbb{R}^{2048}$ from expert demonstrations
$\mathcal{D}_{\text{exp}}$ (50 episodes per task) and train a diffusion-based
policy in two stages: behavioral cloning followed by optional offline RL
fine-tuning.

\textbf{Behavioral Cloning.} We freeze the \methodname and extract
features $f_t \in \mathbb{R}^{2048}$ from expert demonstrations
$\mathcal{D}_{\text{exp}}$. Following DiWA~\cite{chandra2025diwa}, we adopt a
DiffusionMLP policy $\pi_\theta(a_t \mid f_t)$ that generates actions by
iteratively denoising Gaussian noise through $K$ steps:
\begin{equation}
    a_t^{k-1} = \mu_\theta(f_t, a_t^k, k) + \sigma_k \epsilon, \quad
    \epsilon \sim \mathcal{N}(0, I),
    \label{eq:denoise}
\end{equation}
where $a_t^K \sim \mathcal{N}(0, I)$ and $a_t^0$ are the predicted actions.
The policy is trained by minimizing the denoising objective:
\begin{equation}
    \mathcal{L}_{\text{BC}} = \mathbb{E}_{k, \epsilon, (f_t, a_t)}
    \left[\| \mu_\theta(f_t, a_t^k, k) - a_t^{k-1} \|^2 \right].
    \label{eq:bc_loss}
\end{equation}
The key advantage of WAM is the cascading effect: the action-regularized
encoder produces action-aware $e_t$, which conditions the posterior
$z_t \sim q_\phi(z_t \mid h_t, e_t)$, yielding latent features
$f_t = [h_t; z_t]$ that encode richer action-relevant information. The
diffusion policy trained on these features directly benefits from this
improved representation.

\textbf{Offline Policy Fine-tuning.} After behavioral cloning, we further
improve the policy using DPPO~\cite{Ren2024DiffusionPP} entirely within the frozen
world model's latent space, requiring no physical interactions. During imagined
rollouts, observations are unavailable, so the model relies on the prior
$\hat{z}_t \sim p_\phi(z_t \mid h_t)$ to generate latent states. The cascading
effect of WAM is critical here: the action-regularized encoder shapes the
posterior $z_t$ during training, and the KL loss propagates this action-aware
structure to the prior $\hat{z}_t$, ensuring that imagined rollouts benefit
from action-relevant representations. At each iteration, 50 parallel imagined
rollouts are generated within $\mathcal{M}_{\text{wm}}$ using the current
policy. The fine-tuning objective maximizes:
\begin{equation}
    \theta^* = \arg\max_\theta\;
    \mathbb{E}_{\tau \sim \pi_\theta, P_\phi}
    \left[\sum_{t=0}^{T} \gamma^t R_\psi(z_t, a_t)\right],
    \label{eq:finetune}
\end{equation}
where $R_\psi$ is a binary reward classifier. Since WAM produces different
latent representations than the baseline DreamerV2, we retrain the reward
classifiers on WAM features following DiWA's pipeline~\cite{chandra2025diwa}.
All 8 classifiers achieve ${\geq}$0.97 precision and 1.00 recall, confirming
reliable reward estimation in WAM's latent space. The policy is updated via
clipped PPO. During rollouts, the diffusion policy uses 10 denoising
steps with BC regularization ($\alpha_{\text{BC}}\!=\!0.025$) to prevent
catastrophic forgetting. As illustrated in
Fig.~\ref{fig:rssm_unrolled}, the action-aware structure propagates from the
encoder through the prior to the imagined rollouts, benefiting both policy
optimization and reward estimation during offline adaptation.

\section{Experiments}

In our experiments, we aim to study the following questions:
\begin{itemize}
    \item Does the action regularization strategy effectively enhance the performance of \methodname compared to the regular world model?
    \item Does \methodname, with its action-regularized representations, improve downstream policy learning in both offline behavioral cloning and online RL fine-tuning?
\end{itemize}

\vspace{-6pt}
\begin{figure*}[t]
\centering
\includegraphics[width=\textwidth]{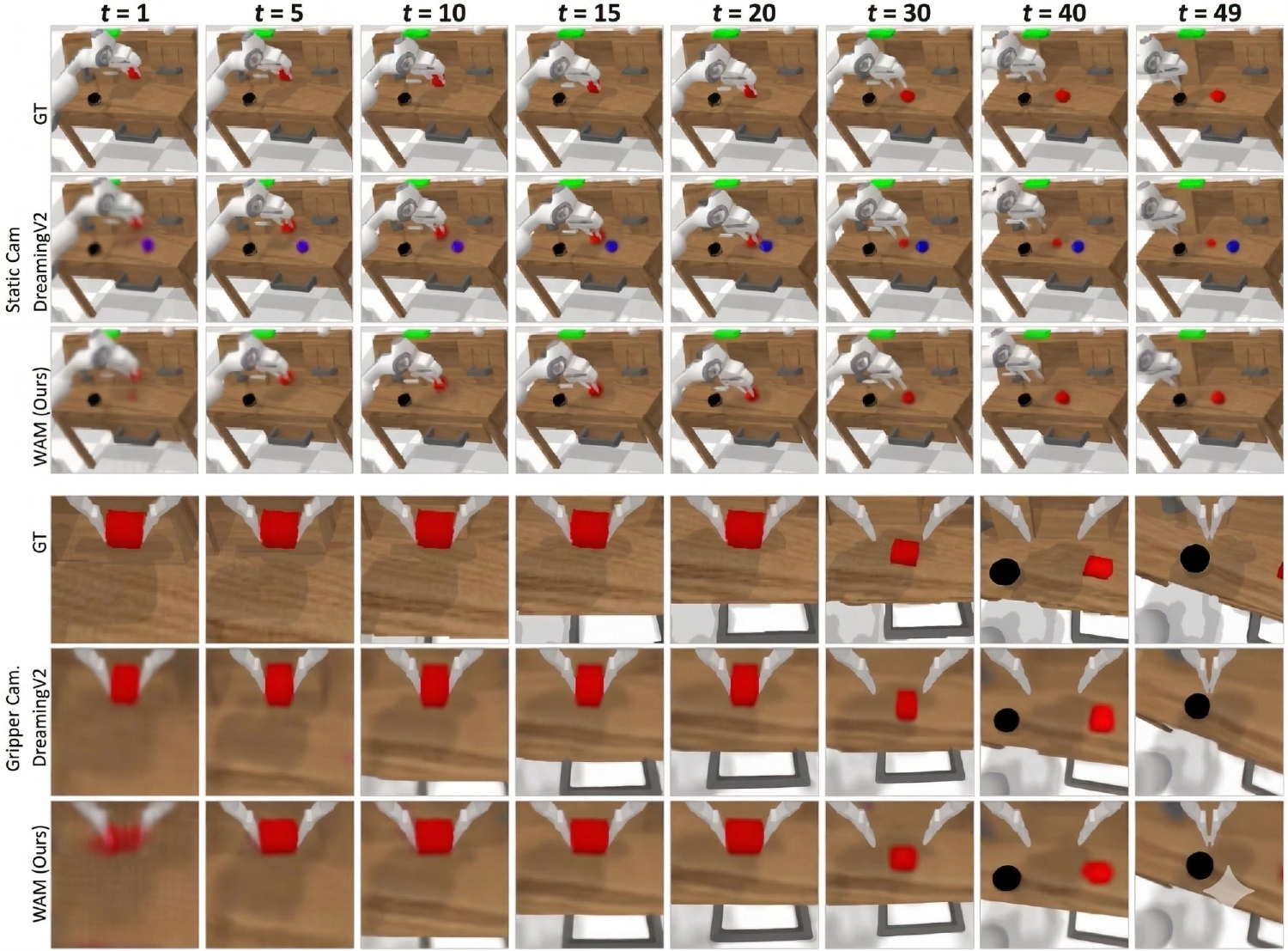}
\caption{\textbf{Qualitative comparison of imagined rollouts on the CALVIN benchmark.} We visualize predicted frames at selected timesteps from both static and gripper cameras. Compared to DreamerV2, our WAM produces more realistic future state predictions across the entire rollout horizon.}
\label{fig:qualitative}
\end{figure*}

 We conduct experiments on \textbf{the CALVIN benchmark}  environment D~\cite{mees2022calvin}, a tabletop manipulation benchmark with a 7-DoF Franka Emika Panda robot. The benchmark provides 6 hours of teleoperated play data ($\sim$500K transitions). Observations consist of $64 \times 64$ RGB images from static and gripper cameras. We train WAM on the play data and evaluate it on eight manipulation tasks.

\subsection{Action Regularization Enhances WAM.}

\subsubsection{\textbf{Baselines}} We assess generation quality by comparing \methodname against the baseline world model DreamerV2~\cite{okada2022dreamingv2}. We sample 100 random sequences from the CALVIN validation set and perform 50-step open-loop imagination rollouts, conditioned on the first ground-truth observation and using ground-truth actions. We report PSNR, SSIM, LPIPS~\cite{Zhang_Isola_Efros_Shechtman_Wang_2018}, and FVD~\cite{Unterthiner2018TowardsAG} to evaluate generation quality.

\subsubsection{\textbf{Training Settings}} We use a sequence length of $T = 50$, a batch size of 500, and a learning rate of $3 \times 10^{-4}$ with AdamW (weight decay $0.05$). The KL balance coefficient is set to $\alpha=0.8$, with loss weights $\lambda_{\text{KL}}=3.0$, $\lambda_{\text{img}}=1.0$, and $\lambda_{\text{act}}=1000.0$. The action prediction loss uses L1 regression. WAM is trained for only 230K gradient steps on $\sim$512K play frames from the CALVIN dataset, roughly $8.7\times$ fewer than the 2M steps used by the baseline DreamingV2, while still achieving stronger downstream policy performance.
 
\begin{table}[t]
\centering
\caption{\textbf{Quantitative evaluation of imagination quality on the CALVIN benchmark.} We report standard video prediction metrics between our World-Action Model (WAM) and the baseline (DreamerV2).}
\label{tab:imagination_quality}
\resizebox{0.45\textwidth}{!}{
\begin{tabular}{lcc}
\toprule
Metric & Ours (WAM) & Baseline (DreamerV2)\\
\midrule
PSNR $\uparrow$    & \textbf{22.10 $\pm$ 2.22}  & 21.66 $\pm$ 2.20 \\
SSIM $\uparrow$    & \textbf{0.814 $\pm$ 0.061} & 0.807 $\pm$ 0.067 \\
LPIPS $\downarrow$ &  \textbf{0.144 $\pm$ 0.072} & 0.149 $\pm$ 0.073 \\
FVD $\downarrow$   & \textbf{10.82}    & 12.13 \\
\bottomrule
\end{tabular}}
\end{table}

\subsubsection{\textbf{Results}} 
Table~\ref{tab:imagination_quality} presents the quantitative evaluation of generation quality on the CALVIN benchmark. Our World-Action Model consistently outperforms the DreamingV2 baseline across all four metrics (PSNR, SSIM, LPIPS, FVD). These improvements stem from the inverse dynamics head added during training, which regularizes the encoder to retain action-relevant information in its latent representations. By requiring the model to predict the action that caused each state transition, the encoder learns to capture not only visual appearance but also the causal structure underlying the observations, leading to more faithful future state generation.

Figure~\ref{fig:qualitative} provides a qualitative comparison of generated rollouts from both the static and gripper cameras. Compared to DreamingV2, our \methodname produces more realistic future state predictions across the entire rollout horizon, with better preservation of object shapes, finer details, and more accurate colors, while DreamingV2 shows noticeable color drift and distorted object shapes.

\subsection{Enhancing Policy Learning with \methodname}

\subsubsection{\textbf{Baselines}} We evaluate \methodname on downstream policy learning by comparing against DiWA~\cite{chandra2025diwa}. As summarized in Table~\ref{tab:method_comparison}, both methods follow the same pipeline of world model pretraining, behavioral cloning, and optional online fine-tuning, with an identical policy architecture (DiffusionMLP). We report behavioral cloning success rates and online RL fine-tuning efficiency.
\begin{table}[t]
\centering
\caption{\textbf{Baseline comparison.} For downstream policy learning, we share the same policy architecture and training pipeline as DiWA, differing only in the world model.}
\label{tab:method_comparison}
\resizebox{0.45\textwidth}{!}{
\begin{tabular}{lcc}
\toprule
 & DiWA & Ours \\
\midrule
World Model          & DreamerV2 & WAM \\
Diffusion Policy     & DiffusionMLP & DiffusionMLP \\
\bottomrule
\end{tabular}}
\end{table}

\begin{table}[t]
\centering
\caption{Behavioral Cloning (BC) success rates (\%) on 8 CALVIN manipulation tasks. Both methods
use identical diffusion policy architectures and training procedures; the only
difference is the world model used for feature extraction.}
\label{tab:bc_results}
\resizebox{0.45\textwidth}{!}{
\begin{tabular}{lcc}
\toprule
\textbf{Task} & \textbf{DiWA} & \textbf{WAM (Ours)} \\
\midrule
close\_drawer        & $58.6 \pm 4.2$ & $\mathbf{89.7 \pm 3.1}$ \\
open\_drawer         & $53.3 \pm 5.1$ & $\mathbf{73.3 \pm 4.8}$ \\
move\_slider\_left   & $50.0 \pm 3.7$ & $\mathbf{68.8 \pm 5.2}$ \\
move\_slider\_right  & $51.7 \pm 4.5$ & $\mathbf{82.8 \pm 3.9}$ \\
turn\_on\_lightbulb  & $42.4 \pm 3.3$ & $\mathbf{51.5 \pm 4.6}$ \\
turn\_off\_lightbulb &  $3.4 \pm 1.8$ & $\mathbf{17.2 \pm 3.4}$ \\
turn\_on\_led        & $\mathbf{44.8 \pm 3.9}$ & $41.4 \pm 4.1$ \\
turn\_off\_led       & $62.5 \pm 5.3$ & $\mathbf{68.8 \pm 4.7}$ \\
\midrule
\textbf{Average}     & $45.8$ & $\mathbf{61.7}$ \\
\bottomrule
\end{tabular}}
\vspace{-0.3cm}
\end{table}

\subsubsection{\textbf{Behavioral Cloning Training Settings}}

Following the DiWA pipeline, we first pretrain the diffusion policy via behavioral cloning. Both DiWA and our method use the same frozen encoder and RSSM to extract $f_t \in \mathbb{R}^{2048}$ features from 50 expert demonstrations per task. The downstream policy is a DiffusionMLP with $K\!=\!20$ denoising steps and action horizon $T_a\!=\!4$, trained for 5{,}000 epochs with a batch size of 256, a learning rate of $10^{-4}$ (cosine decay to $10^{-5}$), weight decay of $10^{-6}$, and EMA with a decay of 0.995. The only difference between the two conditions is which world model provides the features.

\begin{table*}[t]
\centering
\caption{Diffusion policy success rates (\%) on 8 CALVIN tasks after 800 iterations of model-based PPO fine-tuning. DPPO columns report environment steps required to match DiWA performance (lower = more sample-efficient).}
\label{tab:finetune}
\setlength{\tabcolsep}{4pt}
\renewcommand{\arraystretch}{0.9}
\resizebox{\textwidth}{!}{%
\large
\begin{tabular}{l|c|c|c|c|c}
\toprule
\multirow{2}{*}{\textbf{Task}}
  & \textbf{Diffusion Policy}
  & \textbf{DiWA}
  & \textbf{WAM (Ours)}
  & \multicolumn{2}{c}{\textbf{DPPO}} \\
\cmidrule(lr){2-2}\cmidrule(lr){3-3}\cmidrule(lr){4-4}\cmidrule(lr){5-6}
  & Base
  & Offline Fine-Tuning
  & Online Fine-Tuning
  & \makecell{Vision WM Encoder \\ Env Steps to Match DiWA}
  & \makecell{Vision \\ Env Steps to Match DiWA} \\
\midrule
open drawer         & 73.3 $\pm$ 4.8  & 74.44 $\pm$ 1.92          & \textbf{96.7 $\pm$ 2.4}  & 117,600 $\pm$ 23,758   & 134,400 $\pm$ 26,508    \\
close drawer        & 89.7 $\pm$ 3.1  & 91.95 $\pm$ 1.99          & \textbf{96.6 $\pm$ 1.8}  & 600,600 $\pm$ 27,651   & 1,545,600 $\pm$ 261,346 \\
move slider left    & 68.8 $\pm$ 5.2  & 83.33 $\pm$ 1.80          & \textbf{87.5 $\pm$ 3.7}  & 270,933 $\pm$ 28,780   & 1,377,600 $\pm$ 251,439 \\
move slider right   & 82.8 $\pm$ 3.9  & 82.76 $\pm$ 3.45          & \textbf{89.7 $\pm$ 3.2}  & 249,600 $\pm$ 09,050   & 537,600 $\pm$ 23,758    \\
turn on lightbulb   & 51.5 $\pm$ 4.6  & 91.92 $\pm$ 1.75          & \textbf{100.0 $\pm$ 0.0} & 302,933 $\pm$ 15,964   & 588,000 $\pm$ 62,859    \\
turn off lightbulb  & 17.2 $\pm$ 3.4  & \textbf{77.01 $\pm$ 1.99} & 75.9 $\pm$ 4.3           & 327,066 $\pm$ 13,546   & 1,260,000 $\pm$ 142,552 \\
turn on LED         & 41.4 $\pm$ 4.1  & 86.21 $\pm$ 3.45          & \textbf{96.6 $\pm$ 2.1}  & 494,933 $\pm$ 45,655   & 2,251,200 $\pm$ 33,940  \\
turn off LED        & 68.8 $\pm$ 4.7  & 82.33 $\pm$ 6.53          & \textbf{100.0 $\pm$ 0.0} & 277,333 $\pm$ 31,928   & 184,800 $\pm$ 23,758    \\
\midrule
\textbf{Total Physical Interactions} & - & 0 & 0 & ${\sim}$2.5M & ${\sim}$8M \\
\bottomrule
\end{tabular}%
}
\end{table*}


\subsubsection{\textbf{Online Policy Fine-tuning Settings}}
Following DiWA~\cite{chandra2025diwa}, we fine-tune the BC-pretrained diffusion policies using DPPO~\cite{Ren2024DiffusionPP} entirely within the frozen world model's latent space, requiring no physical interactions during policy optimization. At each iteration, the current policy generates 50 parallel imagined rollouts within the frozen world model's latent space. The policy is updated via a clipped PPO surrogate objective with a batch size of 7{,}500, 10 update epochs per iteration, actor learning rate $10^{-5}$, critic learning rate $10^{-3}$, $\gamma = 0.999$, and GAE $\lambda = 0.95$. During rollouts, the diffusion policy uses 10 denoising steps with BC regularization ($\alpha_{\text{BC}} = 0.025$) to prevent catastrophic forgetting. Fine-tuning runs for 800 iterations with evaluation every 25 iterations.

Crucially, since the WAM encoder produces different latent representations than the baseline DreamerV2, we retrain the reward classifiers on WAM features following DiWA's full pipeline: (1)~featurize the entire CALVIN play dataset with the WAM encoder, (2)~extract matched features and raw states from the same 50 expert episodes per task (seed 42), (3)~generate ground-truth reward labels by replaying expert actions in the CALVIN simulator, (4)~augment training data with imagined rollouts inside the WAM, (5)~balance class distribution, and (6)~train contrastive reward classifiers. All 8 classifiers achieve ${\geq}$0.97 precision and 1.00 recall on training data, confirming reliable reward estimation in WAM's latent space.

\subsubsection{\textbf{Behavioral Cloning Results}}
Each policy is evaluated on 29 held-out initial configurations per task.
The frozen world model encodes the current observation into $f_t$, and
the policy generates a 4-step action chunk via DDPM sampling (up to 18
decision points per episode, 72 max steps). An episode succeeds if
CALVIN's built-in task checker confirms completion.
Table~\ref{tab:bc_results} reports results across all eight CALVIN tasks.
WAM outperforms the DiWA baseline on 7 out of 8 tasks, achieving an
average success rate of 61.7\% compared to 45.8\%. The largest gains
appear on tasks involving articulated objects: \textit{close\_drawer}
(+31.1pp), \textit{move\_slider\_right} (+31.1pp), and
\textit{open\_drawer} (+20.0pp), where precise position control is
critical, and the encoder must capture fine-grained spatial cues that
distinguish successful manipulation trajectories. The only task where the
baseline marginally outperforms WAM is \textit{turn\_on\_led} (44.8\%
vs.\ 41.4\%), which we attribute to evaluation variance given the small
number of test episodes.
These results confirm that by jointly training on action prediction and state prediction, WAM learns representations that are more informative for downstream imitation learning without any modification to the policy architecture or training procedure.

\begin{figure*}[t]
    \centering
    \includegraphics[width=\linewidth]{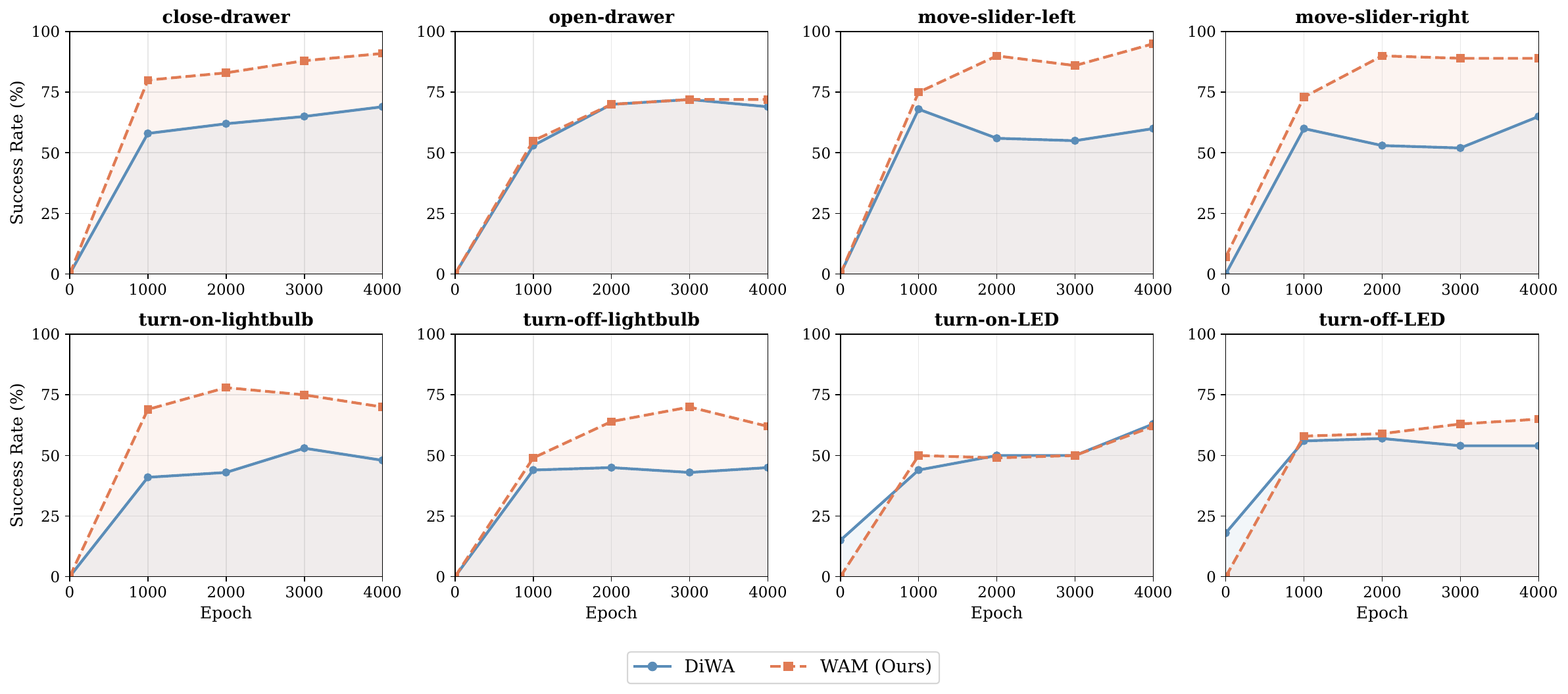}
    \caption{Behavioral cloning evaluation curves across all eight CALVIN tasks. WAM (orange) consistently reaches higher success rates than DiWA (blue) and converges faster on most tasks.}
    \label{fig:bc_curves}
\end{figure*}

Figure~\ref{fig:bc_curves} shows the evaluation curves during BC training across all eight tasks. WAM consistently reaches higher success rates than DiWA and converges faster on most tasks.

\vspace{-6pt}

\subsubsection{\textbf{Diffusion Policy Fine-tuning Results}}
Table~\ref{tab:finetune} reports the results. After 800 iterations of fine-tuning, WAM reaches a 92.8\% average success rate across all eight tasks, outperforming DiWA (79.8\%) by 13.0 percentage points. WAM outperforms DiWA on every task, with
\textit{turn\_on\_lightbulb} and \textit{turn\_off\_led} reaching 100\% success.
The largest gains appear on \textit{open\_drawer} (96.7\% vs.\ 70.0\%) and
\textit{turn\_on\_led} (96.6\% vs.\ 86.2\%), suggesting that action-regularized
representations provide a better optimization landscape for policy fine-tuning
within the world model. Combined with the BC results, these findings demonstrate
that WAM features benefit both offline imitation and online policy adaptation.

These results show that jointly training on action and state prediction not only improves BC policy quality (61.7\% vs.\ 45.8\% average) but also provides a stronger starting point for model-based RL fine-tuning (92.8\% vs.\ 79.8\% average), yielding a 13.0 percentage point gain over DiWA with the same policy architecture and $8.7\times$ fewer world model training steps.

\section{Conclusion}
\label{sec:conclusion}

We presented the World-Action Model (\methodname), a world model augmented with an inverse dynamics objective that regularizes learned representations toward action-relevant structure. The key insight is that under standard reconstruction objectives, the latent state $z_t$ fed to the downstream diffusion policy is never explicitly trained to be action-informative. By jointly predicting actions and future states, \methodname\ propagates action-relevant gradients through the representation pathway, improving the quality of $z_t$ for downstream control. Experiments on eight CALVIN manipulation tasks show that \methodname\ improves behavioral cloning success from 45.8\% to 61.7\% and model-based RL fine-tuning from 79.8\% to 92.8\%, while using $8.7\times$ fewer world model training steps than the baseline.

\bibliographystyle{IEEEtran}
\bibliography{main}

\end{document}